\documentclass[10pt,twocolumn,letterpaper]{article}

\usepackage{cvpr}              

\usepackage{graphicx}
\usepackage{amsmath}
\usepackage{amssymb}
\usepackage{booktabs}

\usepackage[pagebackref,breaklinks,colorlinks]{hyperref}

\usepackage[capitalize]{cleveref}
\crefname{section}{Sec.}{Secs.}
\Crefname{section}{Section}{Sections}
\Crefname{table}{Table}{Tables}
\crefname{table}{Tab.}{Tabs.}

\def\cvprPaperID{*****} 
\def\confName{CVPR}
\def\confYear{2022}

\begin{document}

\title{An Objective Method for Pedestrian Occlusion Level Classification}

\author{Shane Gilroy \textsuperscript{1,2}, Martin Glavin\textsuperscript{1}, Edward Jones\textsuperscript{1} and Darragh Mullins\textsuperscript{1}\\
\textsuperscript{1}National University of Ireland, Galway,
Ireland \\
\textsuperscript{2}Atlantic Technological University,
Ireland\\
}

\maketitle

\begin{abstract}
   
    Pedestrian detection is among the most safety-critical features of driver assistance systems for autonomous vehicles. One of the most complex detection challenges is that of partial occlusion, where a target object is only partially available to the sensor due to obstruction by another foreground object. A number of current pedestrian detection benchmarks provide annotation for partial occlusion to assess algorithm performance in these scenarios, however each benchmark varies greatly in their definition of the occurrence and severity of occlusion. In addition, current occlusion level annotation methods contain a high degree of subjectivity by the human annotator. This can lead to inaccurate or inconsistent reporting of an algorithm's detection performance for partially occluded pedestrians, depending on which benchmark is used. This research presents a novel, objective method for pedestrian occlusion level classification for ground truth annotation. Occlusion level classification is achieved through the identification of visible pedestrian keypoints and through the use of a novel, effective method of 2D body surface area estimation. Experimental results demonstrate that the proposed method reflects the pixel-wise occlusion level of pedestrians in images and is effective for all forms of occlusion, including challenging edge cases such as self-occlusion, truncation and inter-occluding pedestrians.

\end{abstract}


\section{Introduction}

Robust pedestrian detection is one of the most safety-critical features of driver assistance systems and autonomous vehicles. Pedestrian detection is particularly challenging due to the deformable nature and irregular profile of the human body in motion and the inconsistency of color information due to clothing, that can enhance or camouflage any part of a pedestrian. Pedestrian detection systems have improved significantly in recent years with the proliferation of deep learning based solutions and the availability of larger and more diverse datasets. Despite this, many challenges still exist before we reach the detection capabilities required for safe autonomous driving. One of the most complex scenarios is that of partial occlusion, where a target object is only partially available to the sensor due to obstruction by another foreground object. The frequency and variety of occlusion in the automotive environment is substantial and is impacted by both natural and man-made infrastructure as well as the presence of other road users \cite{zhang2020feature}\cite{ruan2021occluded}\cite{vebjorn2021illusion}. Pedestrians can be occluded by static or dynamic objects, may inter-occlude (occlude one another) such as in crowds, and self-occlude - where parts of a pedestrian overlap.
State of the art pedestrian detection solutions claim a detection performance of approximately 65\%-75\% of partially and heavily occluded pedestrians respectively using current benchmarks \cite{gilroy2019overcoming}\cite{ning2021survey}\cite{cao2021handcrafted}\cite{xiao2021deep}. However, the definition of the occurrence and severity of occlusion varies greatly, and a high degree of subjectivity is used to categorize pedestrian occlusion level in each benchmark as shown in Table 1. In addition to this, occurrences of self occlusion, where one part of the body occludes another, has typically been overlooked entirely when categorizing occlusion level. This can lead to inaccurate or inconsistent reporting of a pedestrian detection algorithm’s performance, depending on which dataset is used to verify detection performance \cite{gilroy2019overcoming}\cite{hasan2021generalizable}. In order to address this issue, a universal metric and an objective, repeatable method of occlusion level classification is required for ground truth annotation so that algorithms can be evaluated and compared on an equal scale. 

\begin{table}[]
\caption{Categories of occlusion levels by dataset.}
\begin{tabular}{|l|c|c|c|}
\hline
\textbf{Dataset} & \multicolumn{3}{c|}{\textbf{Occlusion Level}} \\ \cline{2-4} 
 & \textit{Low} & \textit{Partial} & \textit{Heavy} \\ \hline
\begin{tabular}[c]{@{}l@{}}EuroCity\\ Persons \cite{braun2019eurocity}\end{tabular} & \textless{}40\% & 40-80\% & \textgreater{}80\% \\ \hline
CityPersons \cite{zhang2017citypersons} & - & \textless{}35\% & 35-75\% \\ \hline
KITTI \cite{geiger2012we} & \begin{tabular}[c]{@{}c@{}}"Fully \\ Visible"\end{tabular} & \begin{tabular}[c]{@{}c@{}}"Partially \\ Occluded"\end{tabular} & \begin{tabular}[c]{@{}c@{}}"Difficult \\ to See"\end{tabular} \\ \hline
\begin{tabular}[c]{@{}l@{}}Caltech \\ Pedestrian \cite{dollar2009pedestrian}\end{tabular} & - & 1-35\% & 35-80\% \\ \hline
\begin{tabular}[c]{@{}l@{}}Multispectral \\ Pedestrian \cite{hwang2015multispectral}, \\ OVIS \cite{qi2021occluded}\end{tabular} & - & $\leq${}50\% & \textgreater{}50\% \\ \hline
TJU-DHD \cite{pang2020tju} & - & $\leq${}35\% & \textgreater{}35\% \\ \hline
\begin{tabular}[c]{@{}l@{}}Daimler \\ Tsinghua \cite{li2016new}\end{tabular} & \textless{}10\% & 10-40\% & 41-80\% \\ \hline
Li \textit{et al} 2017 \cite{LiUni2017} & \begin{tabular}[c]{@{}c@{}}"Fully \\ Visible"\end{tabular} & 1-40\% & 41-80\% \\ \hline

SAIL-VOS \cite{hu2019sail} & - & 1-25\% & \textgreater{}25-75\% \\ \hline

\hline

\end{tabular}%
\end{table}




This research proposes a novel, objective and consistent method for pedestrian occlusion level classification for ground truth annotation of partially occluded pedestrians. The proposed method more accurately represents the pixel-wise occlusion level than the current state of the art and works for all forms of occlusion including challenging edge cases such as self-occlusion, inter-occluding pedestrians and truncation. 

The contributions of this research are threefold:
1. A novel, objective method for pedestrian occlusion level classification for ground truth annotation is presented.
2. A novel method for estimating the visible 2D body surface area of pedestrians in images.
3. The proposed method is the first occlusion level classifier to infer the level of pedestrian self-occlusion.



\section{Related Work}

This section provides an overview of current occlusion level classification methods for pedestrian detection, pedestrian occlusion level analysis for flood level assessment and commonly used methods for estimating total body surface area.

A number of publicly available datasets provide annotation of the level of pedestrian occlusion in the automotive environment. Table 1 provides an overview of the categories used to define the severity of occlusion in current popular datasets. Analysis of current benchmarks demonstrate the range of inconsistency and subjectivity in the definition of low, partial and heavy occlusion. 
The Eurocity Persons Dataset \cite{braun2019eurocity} categorizes occlusion into three distinct levels: low occlusion (10\%-40\%), moderate occlusion (40\%-80\%), and strong occlusion (larger than 80\%). Classification is carried out by human annotators. The full extent of the occluded pedestrian is estimated, and the approximate level of occlusion is then estimated to be within one of the three defined categories. This process is also used to classify the level of truncation of pedestrians near the image border. A similar approach is undertaken in the Caltech Pedestrian \cite{dollar2009pedestrian}\cite{zhang2016far}, TJU-DHD-pedestrian \cite{pang2020tju}, CrowdHuman \cite{shao2018crowdhuman} and PedHunter \cite{chi2020pedhunter} datasets in which pedestrians are annotated with two bounding boxes that denote the visible and full pedestrian extent. In the case of occluded pedestrians, the location of hidden parts of the full pedestrian were estimated by the human annotator in order to calculate the occlusion ratio. 
Further analysis of the Caltech Pedestrian \cite{dollar2009pedestrian} dataset determined that the probability of occlusion in the automotive environment is not uniform, but rather has a strong bias for the lower portion of the pedestrian to be occluded and for the top portion to be visible. 
Classification of occluded pedestrians in the CityPersons dataset \cite{zhang2017citypersons} is achieved by drawing a line from the top of the head to the middle of the two feet of the occluded pedestrian. Human annotators are required to estimate the location of the head and feet if these are not visible. A bounding box (“$BB-full$”) is then generated for the full pedestrian area using a fixed aspect ratio of 0.41(width/height). A visible pedestrian area bounding box (“$BB-vis$”) is also annotated and the occlusion ratio is calculated as $Area(BB-vis)/Area(BB-full)$. These estimates of occlusion level are then categorized into two levels in the Citypersons benchmark, Reasonable ($<$=35\% occluded) and Heavy Occlusion (35\%-75\%). 
A more semantic approach to determining the occlusion level was taken in the Kitti Vision Benchmark \cite{geiger2012we}, where human annotators were simply asked to mark each bounding box as “visible”, “semi-occluded”, “fully-occluded” or “truncated”. A similar approach was used in the Multispectral Pedestrian Dataset \cite{hwang2015multispectral} where pedestrians "occluded to some extent up to one half" are tagged as partial occlusion; and those whose contour is "mostly occluded" were tagged as heavy occlusion during ground truth annotation.
Occluded Video Instance Segmentation (OVIS) \cite{qi2021occluded} estimates the degree of occlusion by calculating the ratio of intersecting areas of overlapping bounding boxes to the total area of the respective bounding boxes. The authors acknowledge that although this proposed “Bounding Box Occlusion Rate” can be a rough indicator for the degree of occlusion, it can only reflect the occlusion between objects in a partial way and it does not accurately represent the pixel-wise occlusion level of the target objects.

Chaudhary \textit{et al} \cite{chaudhary2019flood}, propose a method of flood level classification from social media images based on the visibility of pedestrians in the image. In this research the average height of a human adult is estimated to be 170cm. The flood level classifier detects pedestrians in an image and estimates how much of the pedestrian is covered by flood water by vertically subdividing the pedestrian into 11 distinct levels. The highest level of the pedestrian occluded by the water indicates the flood height in the image location.
Feng \textit{et al} \cite{feng2020flood} estimates flood level based on the relative height of specific human body parts which are perceived to be below the water line. The water line in the image is hypothesized to be at the bottom line of the bounding box of a person. A similar approach is taken by Quan \textit{et al} \cite{quan2020flood} in which keypoint detection is correlated with a binary mask output of a pedestrian detector. Analysis is then carried out to determine if keypoints which represent the hip or knees are outside of the detected binary mask area due to occlusion by flood water in the image, thereby indicating a relative flood level.
Noh \textit{et al} \cite{noh2018improving} approximate the severity of pedestrian occlusion by dividing a pedestrian bounding box into a 6x3 section grid. Detection confidence values are calculated by applying a pedestrian classifier to grid section and a part confidence map is produced for the complete bounding box.
Zhang \textit{et al} \cite{zhang2018occlusion} assess pedestrian occlusion level by segmenting pedestrians into 5 distinct sections. Each segment is assigned a fixed height and width relative to the total bounding box based on the empirical ratios identified in \cite{felzenszwalb2009object}. ROI pooling is used to detect features within each section and visibility scores are calculated to indicate the relative pedestrian occlusion level.

Wallace \cite{wallace1951exposure} proposed a method of classification of body surface area for the purposes of diagnosing the severity of burn damage of the average adult burn victim \cite{knaysi1968rule}. This method, known as the “Wallace Rule of Nines”, is commonly used by emergency medical providers and first responders to assess the total affected body surface area of burn patients \cite{borhani2017assessment}\cite{tocco2018want}. The Rule of Nines estimates total body surface area by assigning percentages, in multiples of 9\% to semantic body areas, based on the relative physical dimensions of the average adult. The head is estimated to be 9\% of the total body surface area (4.5\% for the front and 4.5\% for the rear). The chest, abdomen, upper back and lower back are each assigned 9\%. Each leg is assigned 18\%, each arm is assigned a total of 9\% and the groin is assigned the remaining 1\%. 
Further research such as \cite{borhani2017assessment}\cite{livingston2000percentage} validate the Rule of Nines for use in the assessment of total body surface area for the average adult, however, provide amendments to more accurately reflect body proportions in specific edge cases such as obese adults and infant children.



\section{Methodology}
An objective method for occlusion level classification is proposed, which removes the subjectivity of the human annotator and more accurately reflects the pixel wise occlusion level than the current state of the art \cite{braun2019eurocity}\cite{zhang2017citypersons}\cite{geiger2012we}\cite{dollar2009pedestrian}\cite{hwang2015multispectral}\cite{qi2021occluded}\cite{li2016new}.
Improving on the concepts originally discussed in \cite{gilroy2021pedestrian}, occlusion level classification consists of 3 steps: 1. Keypoint detection is applied to the input image in order to identify the presence and visibility of specific semantic parts of each pedestrian instance. 2. A visibility threshold is applied and cross-referenced with the pedestrian mask to determine which keypoints are occluded within the image. 3. Visible keypoints are then grouped into larger semantic parts and the total visible surface area is calculated using the 2D body surface area estimation method outlined in Section 3.2 and Figure 1. The proposed method classifies occlusion level for all forms of pedestrian occlusion, including challenging edge cases such as self occlusion, inter-occluding pedestrians and truncation. An overview of the classification pipeline is shown in Figure 2 and qualitative examples of the classifier output for multiple scenarios can be seen in Figure 3.


\subsection{Occluded Keypoint Detection}

Keypoint detection is carried out by a Faster RCNN based keypoint detector using pretrained weights from Detectron2 \cite{wu2019detectron2}. The model uses a ResNet-50-FPN backbone and is trained using the COCO keypoints dataset \cite{lin2014microsoft}.
The keypoint detector outputs 17 keypoints on the human body in addition to a visibility score for each predicted keypoint. Predicted keypoints include shoulders, elbows, wrists, hips, knees and ankles as well as facial characteristics such as nose, eyes and ears. 
A two-step process is then applied to determine the visibility of keypoints in an image. First, a threshold is applied to the keypoint visibility score returned from the keypoint detector. The coordinates of each visible keypoint are then cross-referenced with the pedestrian mask generated by MaskRCNN \cite{he2017mask} to confirm the keypoint location is within the pedestrian mask region in the image. This two-step process increases the identification of occluded keypoints in complex cases such as self-occlusion where the keypoint visibility score is low however the estimated keypoint location may be masked due to the occluding pedestrian region. The presence of specific grouped keypoints indicates the presence of semantic body parts as outlined in Table 2.


\begin{figure}[t]
\begin{center}
    \includegraphics[width=0.42\linewidth]{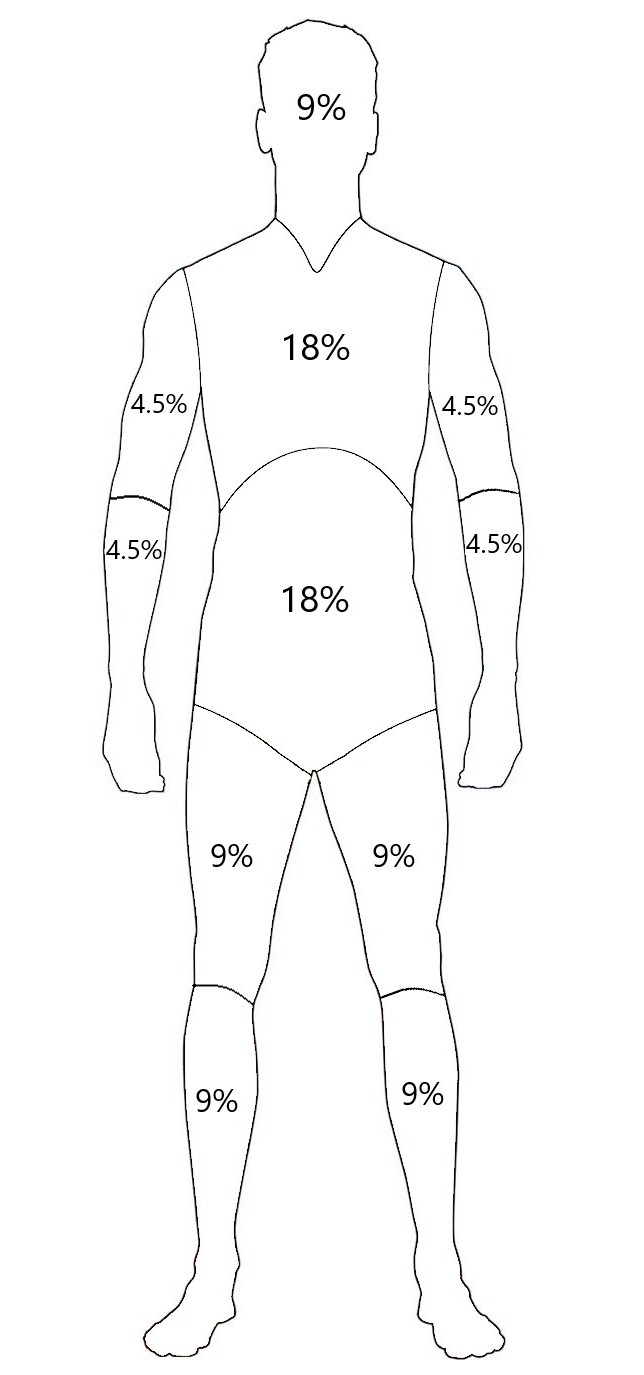}
\end{center}
   \caption{2D Body Surface Area.}
\label{fig:2DBSA}
\label{fig:onecol}
\end{figure}

\begin{table}[]
\caption{Percentage of total body surface area (BSA) and related keypoints for each semantic body part.}
\resizebox{\linewidth}{!}{%
\begin{tabular}{|l|c|}
\hline
\textbf{Body Part (\% BSA)} & \textbf{Related Keypoints} \\ \hline
Head (9\%) & Nose or Eyes or Ears \\ 
Upper Torso (18\%) & Left Shoulder and Right Shoulder  \\ 
Upper Left Arm (4.5\%) & Left Shoulder and Left Elbow  \\ 
Lower Left Arm (4.5\%) & Left Elbow and Left Wrist  \\ 
Upper Right Arm (4.5\%) & Right Shoulder and Right Elbow  \\ 
Lower Right Arm (4.5\%) & Right Elbow and Right Wrist  \\ 
Lower Torso (18\%) & Left Hip and Right Hip  \\ 
Upper Left Leg (9\%) & Left Hip and Left Knee  \\ 
Lower Left Leg (9\%) & Left Knee and Left Ankle  \\
Upper Right Leg (9\%) & Right Hip and Right Knee  \\ 
Lower Right Leg (9\%) & Right Knee and Right Ankle  \\ 

\hline
\end{tabular}
}

\end{table}



\begin{figure*}[t]
\begin{center}
   \includegraphics[width=\textwidth]{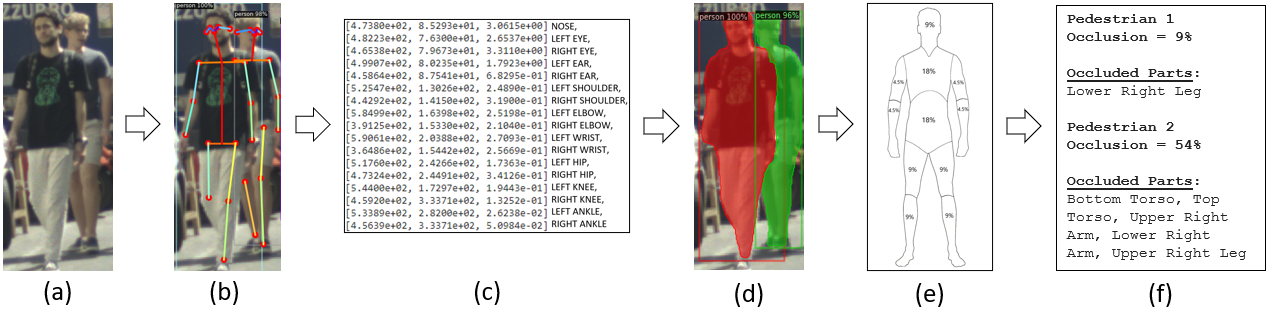}
\end{center}
   \caption{Occlusion level classification overview. (a) Read input image (b) Apply keypoint detection to each pedestrian instance (c) Assess keypoint visibility to identify occluded keypoints (d) Correlate visible keypoints with pedestrian mask to confirm visibility (e) Calculate total visible surface area (f) Output occlusion level classification.}
\label{fig:pipeline}
\end{figure*}


\begin{figure*}
\begin{center}
   \includegraphics[width=\textwidth]{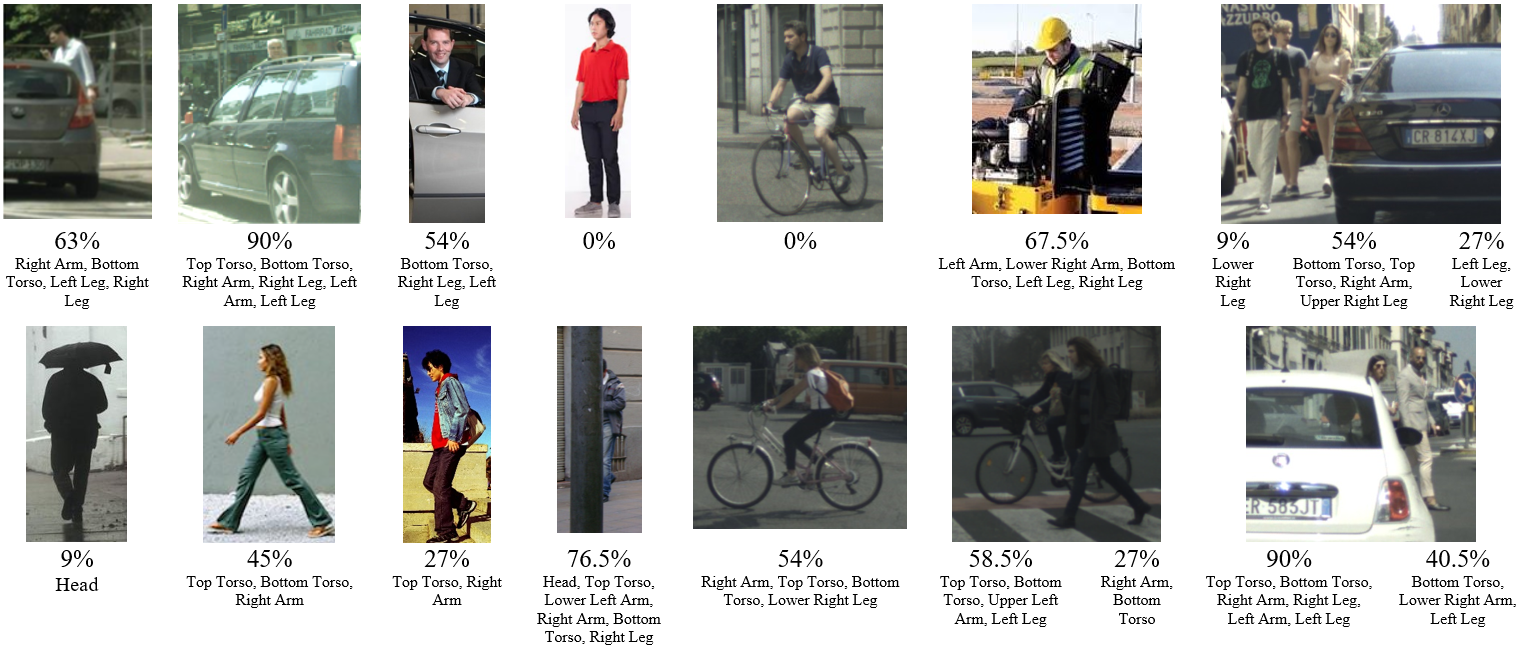}
\end{center}
   \caption{Qualitative validation results. Occlusion level is displayed below each image in addition to a list of occluded semantic parts. Examples are shown for cases of inter-class occlusion, self occlusion and inter-occluding pedestrians. Images containing multiple pedestrian instances read from left to right. All images are compiled from publicly available sources.}
\label{fig:Qual}
\end{figure*}


\subsection{2D Body Surface Area Estimation}

The "Wallace Rule of Nines" \cite{wallace1951exposure} is a time-tested method for determining total body surface area of the average adult. Although effective in the assessment of the body surface area of physical pedestrians, the Rule of Nines is not suitable for assessing the visible surface area of pedestrians in 2D images due to the 3D nature of the human body. An adapted version of the Rule of Nines is proposed for use in determining the visible body surface area of 2D pedestrian images for occlusion level classification. The original proportions of the Rule of Nines have been adjusted respectively to compensate for only one side of the body being visible at any one time, as in the case of 2D images. The proposed method for 2D body surface area estimation is shown in Figure 1. Detected keypoints are related to the semantic body areas in the method shown in Table 2. Examples of the classification output is shown in Figure 3.



\section{Validation}

Qualitative Validation was carried out by applying the proposed method to a wide range of images containing various pedestrian poses, backgrounds and multiple forms of occlusion, including cases of self-occlusion, inter-occluding pedestrians, and truncation. Occlusion level and the occluded semantic parts of each pedestrian instance was deduced using the proposed occlusion level classification method. Human visual inspection was then used to verify the performance of the occlusion level classifier in each case. A custom dataset of 320 images, compiled from multiple publicly available sources including \cite{braun2019eurocity}\cite{zhang2017citypersons}\cite{zhuo2018occluded}\cite{marin2013occlusion}, was used in this validation step to ensure a wide diversity of pedestrian occlusion scenarios. Examples of the qualitative validation are provided in Figure 3.

\subsection{Quantitative Validation}

Quantitative validation was carried out by comparing the proposed method with the calculated pixel-wise occlusion level, derived using MaskRCNN \cite{he2017mask}, and the current state of the art as described in CityPersons \cite{zhang2017citypersons} for both visible and progressively occluded pedestrians.
In order to determine the pixel-wise occlusion, the total pixel area must be calculated for both the fully visible pedestrian and the same pedestrian under occlusion. 
To achieve this, a custom dataset of 200 images was created, including a wide range of occlusion scenarios and challenging pedestrian poses such as walking, running and cycling.  MaskRCNN \cite{he2017mask} was applied to a fully visible reference image and the masked pixel area ($MaskArea_{full}$) was calculated for each pedestrian instance. Occlusions were then superimposed on the reference image and the remaining visible pedestrian pixel area ($Mask Area_{occ}$) is calculated in order to determine the pixel-wise occlusion ratio, Eq.1.

\begin{equation} \label{eq1}
Occ_{pixel} = \frac{Mask Area_{occ}}{Mask Area_{full}} 
\end{equation}

The proposed method was then compared with the pixel-wise occlusion level and the method described in CityPersons \cite{zhang2017citypersons} to determine the pixel-wise accuracy of the proposed occlusion level classifier. More subjective occlusion level classification methods such as those used in \cite{braun2019eurocity}\cite{dollar2009pedestrian}\cite{geiger2012we}\cite{hwang2015multispectral} are omitted for the purposes of this testing. A sample of the images used in these experiments can be seen in Figure 4. Quantitative validation results are provided in Figure 5.




\begin{figure}[t]
\begin{center}
   \includegraphics[width=\linewidth]{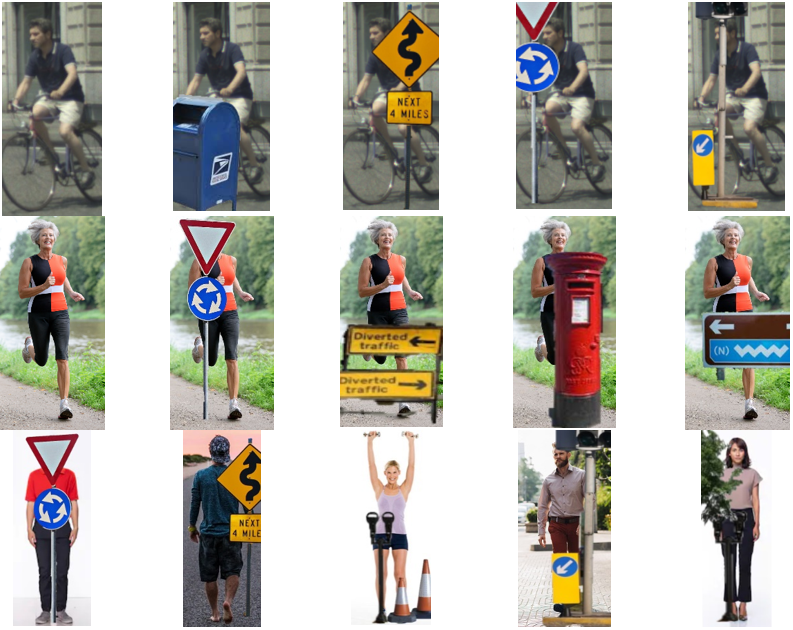}
\end{center}
   \caption{Quantitative validation dataset sample images. The custom dataset consists of 200 images covering a wide range of pedestrian poses and superimposed occlusions. All images are compiled from publicly available sources.}
\label{fig:QuantDS}
\end{figure}




\begin{figure}[t]
\begin{center}
   \includegraphics[width=\linewidth]{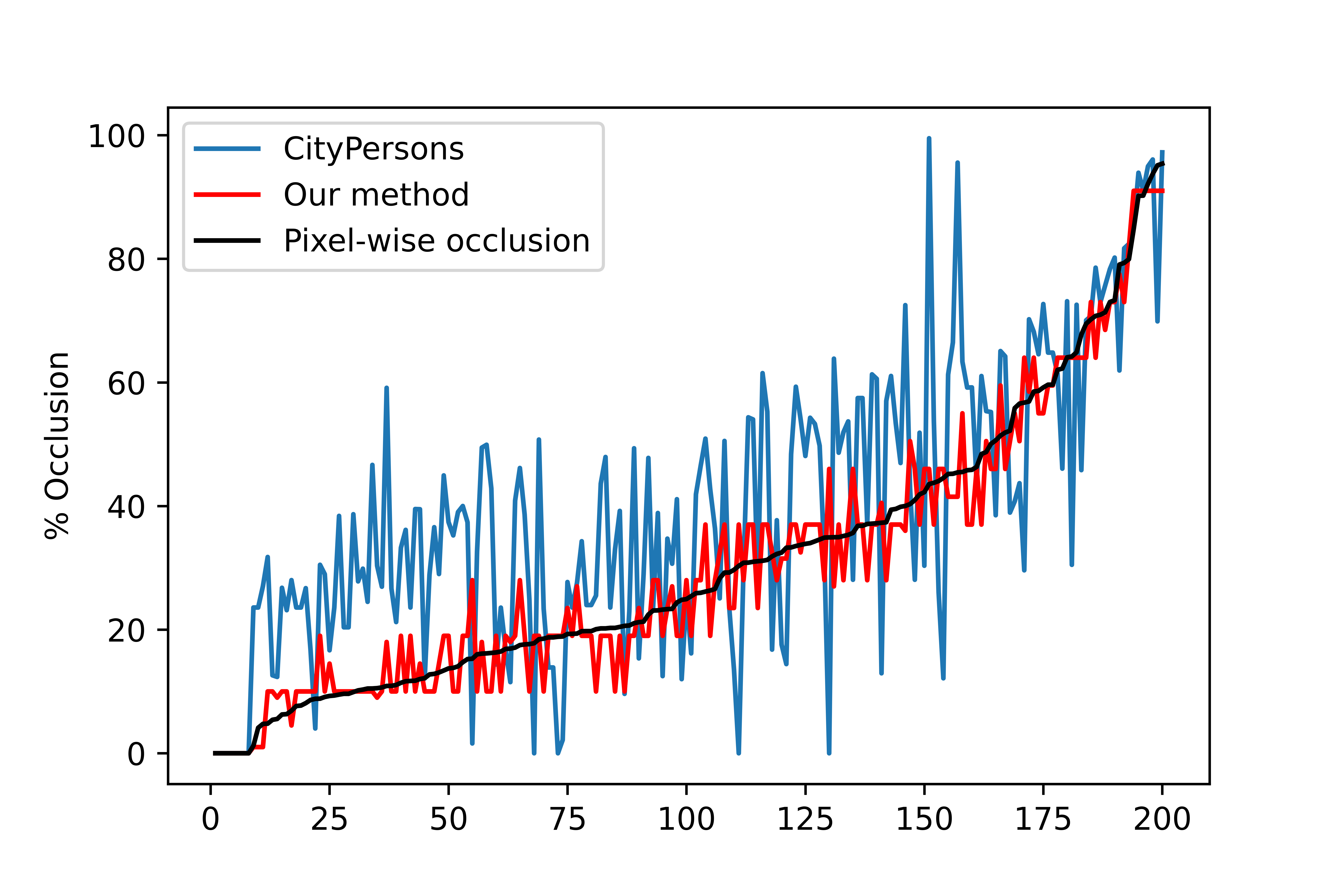}
\end{center}
   \caption{Quantitative Evaluation Results. Our proposed method is compared with the pixel-wise occlusion level as produced by MaskRCNN\cite{he2017mask} and the current state of the art as described in CityPersons\cite{zhang2017citypersons} for a dataset of 200 images. Results demonstrate that our method is a significant improvement over the state of the art when plotted against the pixel-wise occlusion level.}
\label{fig:QuantRES}
\end{figure}



\begin{figure*}[t]
\begin{center}
   \includegraphics[width=.95\textwidth]{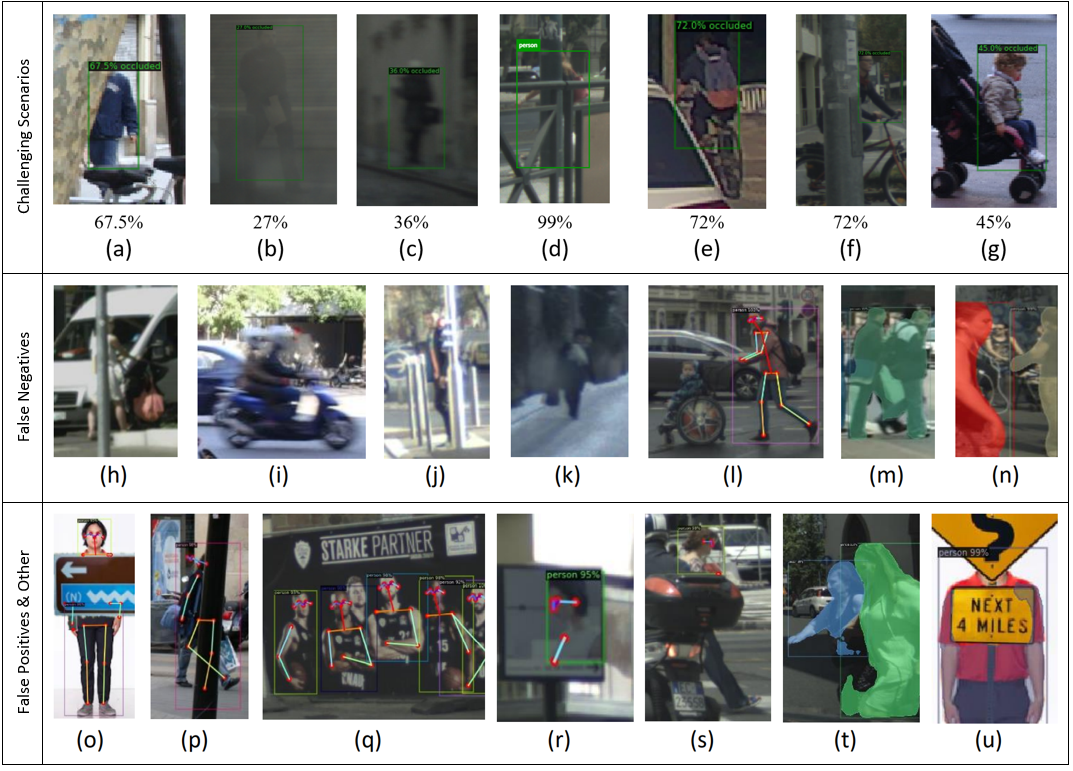}
\end{center}
   \caption{Examples of challenging image frames, false negatives and false positives. The top row provides examples of challenging detection scenarios and displays the occlusion level below each image. The middle row provides examples of false negatives and the bottom row provides examples of false positives and other classification errors.}
\label{fig:Analysis}
\end{figure*}










\section{Discussion and Analysis}
An objective method for occlusion level classification is proposed. The qualitative validation results shown in Figure 3 demonstrate the capability of the proposed method for classifying occlusion level for all forms of occlusion, including challenging edge cases such as self-occlusion, truncation, and inter-occluding pedestrians. By removing the subjectivity of a human annotator, the proposed method is more robust and repeatable than the current state of the art and is suitable for the objective comparison of pedestrian detection algorithms, regardless of the benchmark used.
Classification of pedestrian self-occlusion, heretofore ignored in the assessment of partially occluded pedestrians, may have a large impact on assessing the detectability of pedestrians using modern techniques. This is especially relevant in scenarios where detection confidence is linked to the presence of key salient features which may be self-occluded by the target pedestrian in the image. More detailed analysis of detection performance in cases of self-occlusion will increase our understanding of the behaviour of deep learning-based detection routines. Characterization of detection performance for what were previously considered “visible” pedestrians, in cases where the algorithm specific informative value of a pedestrian is occluded will help identify potential failure modes of current state of the art pedestrian detection systems.

The quantitative validation results shown in Figure 5, demonstrate the proposed method's capability in representing the “real world” or pixel-wise occlusion value for challenging pedestrian poses, regardless of the severity or form of occlusion. The proposed method of 2D body surface area estimation shown in Figure 1, derived from the “Wallace Rule of Nines”, has proven effective in calculating the visible area of partially occluded pedestrians for a wide range of pedestrian poses and occlusion scenarios. Further analysis of the quantitative validation results clearly displays an improvement over the current state of the art \cite{zhang2017citypersons} when compared to the pixel-wise occlusion value. 

\subsection{Challenging Image Frames}

Figure 6 provides a sample of the classifier performance for challenging detection scenarios as well as highlighting classification errors that can occur for indistinct pedestrian instances in particular frames. 
Missed detections or false negatives can occur as a result of low detection confidence of the keypoint detector or MaskRCNN due to excessive motion blur, camera artifacts or low images resolution. Detection confidence is reduced in scenarios where the pedestrian outline closely matches that of the image background. Figure 6 (a), (b) and (c) successfully classify pedestrian occlusion level in cases of heavy occlusion, image glare and low resolution respectively. In each case, the pedestrian outline distinctly differs from the image background. In similar scenarios where the pedestrian outline and the image background are less diverse, such as in Figure 6 (h), (j) and (k), detection confidence is reduced resulting in a false negative.
Keypoint errors can occur in complex detection scenarios which can result in incorrect classification for a particular frame. Occurrences of these have been noted in cases where a pedestrian instance is highly segmented by the occluder, prompting the algorithm to propose multiple pedestrian instances or omitting sections of a pedestrian that appear to be unconnected to the primary pedestrian instance due to intersecting occlusion. Examples of these occurrences can be seen in Figure 6 (o), (p) and (s). 
Similarly, pedestrian mask errors can also occur in challenging frames. Mask errors can include mask leakage, which can falsely indicate the presence of occluded keypoints, Figure 6 (u), and incomplete or imprecise masks which can lead to the false omission of specific keypoints or pedestrian instances as shown in Figure 6 (m), (n) and (t).
Although the proposed method is designed to focus on pedestrians, other road users such as cyclists, motorcyclists and children in strollers may be classified as occluded pedestrians. In addition, person depictions on advertising images and other media may be classified as pedestrians by the algorithm.
Many of the misclassification errors presented can be reduced by further improvement in keypoint and pedestrian mask detection models which can be integrated into the detection pipeline as technology progresses.

\section{Conclusions}
This research proposes an objective method of pedestrian occlusion level classification for ground truth annotation. The proposed method uses keypoint detection and mask segmentation to identify and determine the visibility of the semantic parts of partially occluded pedestrians and calculates a percentage occluded body surface area using a novel, effective method for 2D body surface area estimation. The proposed method removes the subjectivity of the human annotator used by the current state of the art, in turn increasing the robustness and repeatability of pedestrian occlusion level classification. 
Qualitative and quantitative validation demonstrates the effectiveness of the proposed method for all forms of occlusion including challenging edge cases such as self-occlusion and inter-occluding pedestrians. Experimental results show a significant improvement over the current state of the art when plotted against the pixel-wise pedestrian occlusion level.
Universal use of the proposed method will improve the accuracy and consistency of occlusion level annotation in pedestrian detection benchmarks and will improve the precision of occlusion aware pedestrian detection networks. Detailed analysis of edge cases such as self-occlusion, previously overlooked in popular pedestrian detection datasets, will increase our understanding of deep learning-based detection routines, provide more advanced characterisation of pedestrian detection algorithms and help to identify potential failure modes in current technology.

{\small
\bibliographystyle{ieeetr}
\bibliography{egbib}

\begin{thebibliography}{10}

\bibitem{zhang2020feature}
T.~Zhang, Q.~Ye, B.~Zhang, J.~Liu, X.~Zhang, and Q.~Tian, ``Feature calibration
  network for occluded pedestrian detection,'' {\em IEEE Transactions on
  Intelligent Transportation Systems}, 2020.

\bibitem{ruan2021occluded}
B.~Ruan and C.~Zhang, ``Occluded pedestrian detection combined with semantic
  features,'' {\em IET Image Processing}, 2021.

\bibitem{vebjorn2021illusion}
E.~Vebj{\o}rn, S.~Mats, P.~Angelo, B.~Gisela, J.~Sebastian, W.~Johan, H.~Alena,
  {\em et~al.}, ``The illusion of absence: how a common feature of magic shows
  can explain a class of road accidents,'' {\em Cognitive Research}, vol.~6,
  no.~1, 2021.

\bibitem{gilroy2019overcoming}
S.~Gilroy, E.~Jones, and M.~Glavin, ``Overcoming occlusion in the automotive
  environment-a review,'' {\em IEEE Transactions on Intelligent Transportation
  Systems}, 2019.

\bibitem{ning2021survey}
C.~Ning, L.~Menglu, Y.~Hao, S.~Xueping, and L.~Yunhong, ``Survey of pedestrian
  detection with occlusion,'' {\em Complex \& Intelligent Systems}, vol.~7,
  no.~1, pp.~577--587, 2021.

\bibitem{cao2021handcrafted}
J.~Cao, Y.~Pang, J.~Xie, F.~S. Khan, and L.~Shao, ``From handcrafted to deep
  features for pedestrian detection: A survey,'' {\em IEEE Transactions on
  Pattern Analysis and Machine Intelligence}, 2021.

\bibitem{xiao2021deep}
Y.~Xiao, K.~Zhou, G.~Cui, L.~Jia, Z.~Fang, X.~Yang, and Q.~Xia, ``Deep learning
  for occluded and multi-scale pedestrian detection: A review,'' {\em IET Image
  Processing}, vol.~15, no.~2, pp.~286--301, 2021.

\bibitem{hasan2021generalizable}
I.~Hasan, S.~Liao, J.~Li, S.~U. Akram, and L.~Shao, ``Generalizable pedestrian
  detection: The elephant in the room,'' in {\em Proceedings of the IEEE/CVF
  Conference on Computer Vision and Pattern Recognition}, pp.~11328--11337,
  2021.

\bibitem{braun2019eurocity}
M.~Braun, S.~Krebs, F.~Flohr, and D.~M. Gavrila, ``Eurocity persons: A novel
  benchmark for person detection in traffic scenes,'' {\em IEEE transactions on
  pattern analysis and machine intelligence}, vol.~41, no.~8, pp.~1844--1861,
  2019.

\bibitem{zhang2017citypersons}
S.~Zhang, R.~Benenson, and B.~Schiele, ``Citypersons: A diverse dataset for
  pedestrian detection,'' in {\em Proceedings of the IEEE Conference on
  Computer Vision and Pattern Recognition}, pp.~3213--3221, 2017.

\bibitem{geiger2012we}
A.~Geiger, P.~Lenz, and R.~Urtasun, ``Are we ready for autonomous driving? the
  kitti vision benchmark suite,'' in {\em 2012 IEEE Conference on Computer
  Vision and Pattern Recognition}, pp.~3354--3361, IEEE, 2012.

\bibitem{dollar2009pedestrian}
P.~Doll{\'a}r, C.~Wojek, B.~Schiele, and P.~Perona, ``Pedestrian detection: A
  benchmark,'' in {\em 2009 IEEE Conference on Computer Vision and Pattern
  Recognition}, pp.~304--311, IEEE, 2009.

\bibitem{hwang2015multispectral}
S.~Hwang, J.~Park, N.~Kim, Y.~Choi, and I.~So~Kweon, ``Multispectral pedestrian
  detection: Benchmark dataset and baseline,'' in {\em Proceedings of the IEEE
  conference on computer vision and pattern recognition}, pp.~1037--1045, 2015.

\bibitem{qi2021occluded}
J.~Qi, Y.~Gao, X.~Liu, Y.~Hu, X.~Wang, X.~Bai, P.~H. Torr, S.~Belongie,
  A.~Yuille, and S.~Bai, ``Occluded video instance segmentation,'' {\em arXiv
  preprint arXiv:2102.01558}, 2021.

\bibitem{pang2020tju}
Y.~Pang, J.~Cao, Y.~Li, J.~Xie, H.~Sun, and J.~Gong, ``Tju-dhd: A diverse
  high-resolution dataset for object detection,'' {\em IEEE Transactions on
  Image Processing}, vol.~30, pp.~207--219, 2020.

\bibitem{li2016new}
X.~Li, F.~Flohr, Y.~Yang, H.~Xiong, M.~Braun, S.~Pan, K.~Li, and D.~M. Gavrila,
  ``A new benchmark for vision-based cyclist detection,'' in {\em 2016 IEEE
  Intelligent Vehicles Symposium (IV)}, pp.~1028--1033, IEEE, 2016.

\bibitem{LiUni2017}
X.~Li, L.~Li, F.~Flohr, J.~Wang, H.~Xiong, M.~Bernhard, S.~Pan, D.~M. Gavrila,
  and K.~Li, ``A unified framework for concurrent pedestrian and cyclist
  detection,'' {\em IEEE Transactions on Intelligent Transportation Systems},
  vol.~18, no.~2, pp.~269--281, 2017.

\bibitem{hu2019sail}
Y.-T. Hu, H.-S. Chen, K.~Hui, J.-B. Huang, and A.~G. Schwing, ``Sail-vos:
  Semantic amodal instance level video object segmentation-a synthetic dataset
  and baselines,'' in {\em Proceedings of the IEEE/CVF Conference on Computer
  Vision and Pattern Recognition}, pp.~3105--3115, 2019.

\bibitem{zhang2016far}
S.~Zhang, R.~Benenson, M.~Omran, J.~Hosang, and B.~Schiele, ``How far are we
  from solving pedestrian detection?,'' in {\em Proceedings of the iEEE
  conference on computer vision and pattern recognition}, pp.~1259--1267, 2016.

\bibitem{shao2018crowdhuman}
S.~Shao, Z.~Zhao, B.~Li, T.~Xiao, G.~Yu, X.~Zhang, and J.~Sun, ``Crowdhuman: A
  benchmark for detecting human in a crowd,'' {\em arXiv preprint
  arXiv:1805.00123}, 2018.

\bibitem{chi2020pedhunter}
C.~Chi, S.~Zhang, J.~Xing, Z.~Lei, S.~Z. Li, and X.~Zou, ``Pedhunter: Occlusion
  robust pedestrian detector in crowded scenes,'' in {\em Proceedings of the
  AAAI Conference on Artificial Intelligence}, vol.~34, pp.~10639--10646, 2020.

\bibitem{chaudhary2019flood}
P.~Chaudhary, S.~D'Aronco, M.~Moy~de Vitry, J.~P. Leit{\~a}o, and J.~D. Wegner,
  ``Flood-water level estimation from social media images,'' {\em ISPRS Annals
  of the Photogrammetry, Remote Sensing and Spatial Information Sciences},
  vol.~4, no.~2/W5, pp.~5--12, 2019.

\bibitem{feng2020flood}
Y.~Feng, C.~Brenner, and M.~Sester, ``Flood severity mapping from volunteered
  geographic information by interpreting water level from images containing
  people: A case study of hurricane harvey,'' {\em ISPRS Journal of
  Photogrammetry and Remote Sensing}, vol.~169, pp.~301--319, 2020.

\bibitem{quan2020flood}
K.-A.~C. Quan, V.-T. Nguyen, T.-C. Nguyen, T.~V. Nguyen, and M.-T. Tran,
  ``Flood level prediction via human pose estimation from social media
  images,'' in {\em Proceedings of the 2020 International Conference on
  Multimedia Retrieval}, pp.~479--485, 2020.

\bibitem{noh2018improving}
J.~Noh, S.~Lee, B.~Kim, and G.~Kim, ``Improving occlusion and hard negative
  handling for single-stage pedestrian detectors,'' in {\em Proceedings of the
  IEEE Conference on Computer Vision and Pattern Recognition}, pp.~966--974,
  2018.

\bibitem{zhang2018occlusion}
S.~Zhang, L.~Wen, X.~Bian, Z.~Lei, and S.~Z. Li, ``Occlusion-aware r-cnn:
  Detecting pedestrians in a crowd,'' in {\em Proceedings of the European
  Conference on Computer Vision (ECCV)}, pp.~637--653, 2018.

\bibitem{felzenszwalb2009object}
P.~F. Felzenszwalb, R.~B. Girshick, D.~McAllester, and D.~Ramanan, ``Object
  detection with discriminatively trained part-based models,'' {\em IEEE
  transactions on pattern analysis and machine intelligence}, vol.~32, no.~9,
  pp.~1627--1645, 2009.

\bibitem{wallace1951exposure}
A.~Wallace, ``The exposure treatment of burns,'' {\em The Lancet}, vol.~257,
  no.~6653, pp.~501--504, 1951.

\bibitem{knaysi1968rule}
G.~A. KNAYSI, G.~F. CRIKELAIR, and B.~COSMAN, ``The rule of nines: its history
  and accuracy,'' {\em Plastic and reconstructive surgery}, vol.~41, no.~6,
  pp.~560--563, 1968.

\bibitem{borhani2017assessment}
K.~Borhani-Khomani, S.~Partoft, and R.~Holmgaard, ``Assessment of burn size in
  obese adults; a literature review,'' {\em Journal of plastic surgery and hand
  surgery}, vol.~51, no.~6, pp.~375--380, 2017.

\bibitem{tocco2018want}
I.~Tocco-Tussardi, B.~Presman, and F.~Huss, ``Want correct percentage of tbsa
  burned? let a layman do the assessment,'' {\em Journal of Burn Care \&
  Research}, vol.~39, no.~2, pp.~295--301, 2018.

\bibitem{livingston2000percentage}
E.~H. Livingston and S.~Lee, ``Percentage of burned body surface area
  determination in obese and nonobese patients,'' {\em Journal of surgical
  research}, vol.~91, no.~2, pp.~106--110, 2000.

\bibitem{gilroy2021pedestrian}
S.~Gilroy, M.~Glavin, E.~Jones, and D.~Mullins, ``Pedestrian occlusion level
  classification using keypoint detection and 2d body surface area
  estimation,'' in {\em Proceedings of the IEEE/CVF International Conference on
  Computer Vision}, pp.~3833--3839, 2021.

\bibitem{wu2019detectron2}
Y.~Wu, A.~Kirillov, F.~Massa, W.-Y. Lo, and R.~Girshick, ``Detectron2.''
  \url{https://github.com/facebookresearch/detectron2}, 2019.

\bibitem{lin2014microsoft}
T.-Y. Lin, M.~Maire, S.~Belongie, J.~Hays, P.~Perona, D.~Ramanan,
  P.~Doll{\'a}r, and C.~L. Zitnick, ``Microsoft coco: Common objects in
  context,'' in {\em European conference on computer vision}, pp.~740--755,
  Springer, 2014.

\bibitem{he2017mask}
K.~He, G.~Gkioxari, P.~Doll{\'a}r, and R.~Girshick, ``Mask r-cnn,'' in {\em
  Proceedings of the IEEE international conference on computer vision},
  pp.~2961--2969, 2017.

\bibitem{zhuo2018occluded}
J.~Zhuo, Z.~Chen, J.~Lai, and G.~Wang, ``Occluded person re-identification,''
  in {\em 2018 IEEE International Conference on Multimedia and Expo (ICME)},
  pp.~1--6, IEEE, 2018.

\bibitem{marin2013occlusion}
J.~Mar{\'\i}n, D.~V{\'a}zquez, A.~M. L{\'o}pez, J.~Amores, and L.~I. Kuncheva,
  ``Occlusion handling via random subspace classifiers for human detection,''
  {\em IEEE transactions on cybernetics}, vol.~44, no.~3, pp.~342--354, 2013.

\end{thebibliography}
}

\end{document}